\documentclass[10pt]{article} 
\usepackage[preprint]{tmlr}

\usepackage{amsmath,amsfonts,bm}

\def\eqref#1{equation~\ref{#1}}

\def\1{\bm{1}}

\DeclareMathAlphabet{\mathsfit}{\encodingdefault}{\sfdefault}{m}{sl}
\SetMathAlphabet{\mathsfit}{bold}{\encodingdefault}{\sfdefault}{bx}{n}

\usepackage{microtype}
\usepackage{graphicx}
\usepackage{subcaption}
\usepackage{booktabs}
\usepackage{wrapfig}
\usepackage{caption}
\usepackage{float}
\usepackage{hyperref}
\usepackage{ulem}
\usepackage{array}

\usepackage{amsmath}
\usepackage{amssymb}
\usepackage{mathtools}
\usepackage{amsthm}
\usepackage{xcolor}
\usepackage{xspace}
\usepackage{enumitem}
\usepackage{algorithm}
\usepackage{algorithmic}

\usepackage{multirow}
\usepackage{pifont}
\newcommand{\cmark}{\ding{51}}
\newcommand{\xmark}{\ding{55}}

\title{SymMatika: High-Impact Substructure-Reuse for Symbolic Discovery}

\author{%
  Michael Scherk,
  Boyuan Chen\\
  Duke University\\
  \textcolor{orange}{\url{http://www.generalroboticslab.com/SymMatika}}
}

\begin{document}

\maketitle

\begin{abstract}
    Symbolic regression (SR) seeks to recover closed-form mathematical expressions that describe observed data. While existing methods have advanced the discovery of either explicit mappings (i.e., $y = f(\mathbf{x})$) or discovering implicit relations (i.e., $F(\mathbf{x})=0$), few modern and accessible frameworks support both. Moreover, most approaches treat each expression candidate in isolation, without reusing recurring structural patterns that could accelerate search. Those that do reuse substructures select them by frequency among top candidates, which does not measure whether a substructure actually contributes to an expression's fitness. We introduce \textsc{SymMatika}, a symbolic regression framework built around an updating ``motif'' library, which scores subtrees of top-performing candidates by their measured impact on fitness, stores high-impact subtrees in a bounded shared library, and recombines them to guide future generations. The library sits on a feedback-driven evolutionary engine that adapts operator rates and candidate generation to each population's progress, and the same mechanisms support both explicit and implicit relation discovery using implicit-derivative metrics. On Nguyen, \textsc{SymMatika} recovers 96.5\% of expressions on average and 61\% of Nguyen-12, where the next-best method recovers 2\%. On the 119 Feynman equations and 57 SRBench Black-box problems, it lies on the error-complexity Pareto front with the most compact expressions of any competitive method, and it recovers implicit relations on Eureqa's physical systems. These results indicate that impact-based substructure reuse is an effective strategy for scientific discovery. Our complete codebase has been open-sourced.
\end{abstract}

\begin{figure}[H]
    \centering
    \includegraphics[width=\linewidth]{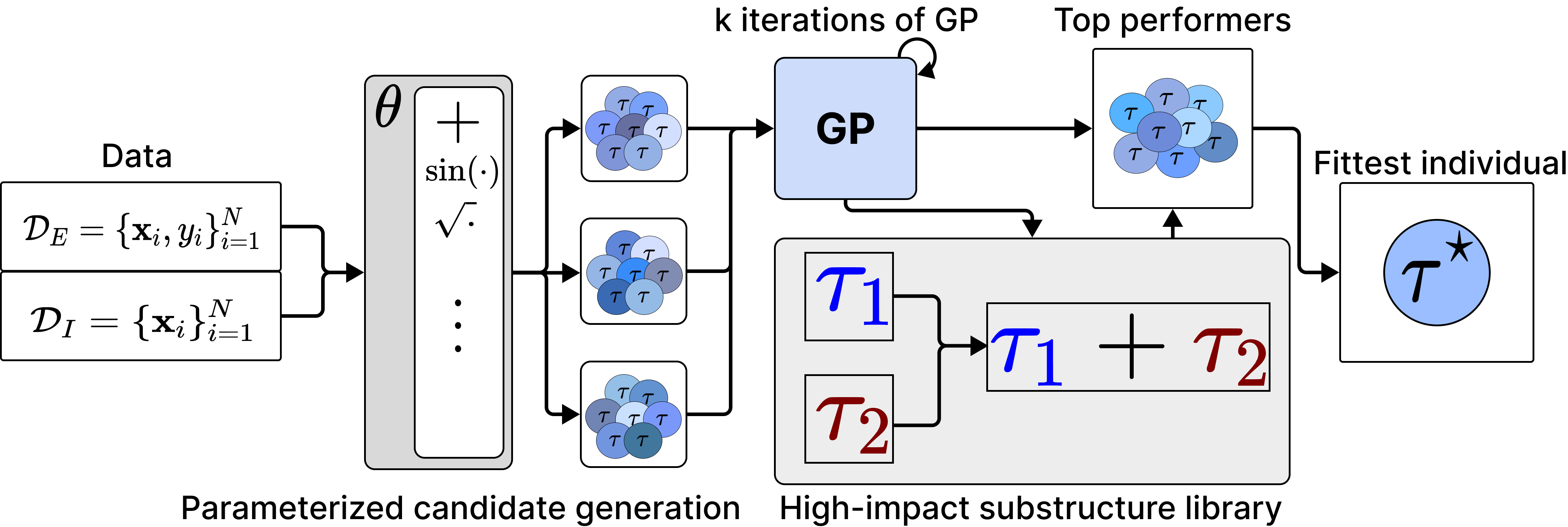}
    \caption{\textsc{SymMatika} is a high-performing symbolic regression framework. Given data $\mathcal{D}_E$ (explicit) or $\mathcal{D}_I$ (implicit), a parameterized candidate generator $\theta$ produces $m$ initial populations. Each population evolves for $k$ GP iterations, and the top-$P$ candidates are used to update both the generator parameters $\theta$ and our reusable substructure library. High-impact substructures are recombined to form new candidates to accelerate convergence. The fittest individual $\tau^\star$ is selected as the final expression.}
    \label{fig:main-model}
    \vspace{-15pt}
\end{figure}

\section{Introduction}

In the late 16\textsuperscript{th} century, Tycho Brahe meticulously recorded the positions of celestial bodies. His data laid the foundation for Johannes Kepler, who in 1601 derived analytical expressions of the motion of these planets. These expressions launched a scientific revolution in their discovery that Mars' orbit was in fact an ellipse. This was an early instance of Symbolic Regression (SR), where governing laws are distilled from observational data as interpretable mathematical expressions. It aims to find these expressions by searching the space of interpretable mathematical expressions. Most methods aim to identify a symbolic function $f: \mathbb{R}^n \rightarrow \mathbb{R}$ that best fits a dataset $(X,y)$, where $X_i \in \mathbb{R}^n$ and $y_i \in \mathbb{R}$. Other methods aim to discover implicit relations $F(\mathbf{x}) = 0$ (e.g. Kepler's equations), with $\mathbf{x} \in \mathbb{R}^n$.

Symbolic regression poses a computationally hard challenge, as the space of symbolic expressions grows exponentially with expression length \citep{lu2016using, virgolin2022symbolic}. The dominant approach, genetic programming (GP) \citep{koza1990genetic}, stochastically evolves populations of candidate expressions using mutation and crossover. Tools such as PySR \citep{cranmer2023interpretable}, Operon \citep{burlacu2020operon}, GP-GOMEA \citep{virgolin2021improving}, and SBP-GP \citep{pawlak2014semantic} apply GP to search across diverse mathematical structures without assuming parametric forms. However, their convergence is often slow due to uninformed exploration, generating many unpromising candidates across generations.

To improve efficiency, neural-guided symbolic regression approaches integrate deep learning to bias the search toward promising expressions. Recent algorithms \citep{petersen2019deep, mundhenk2021symbolic} use Recurrent Neural Networks (RNNs) to learn trends in expression coefficients. However, they do not capture higher-order syntactic structures, such as the recurring mathematical substructures found in top-performing individuals. Therefore, these approaches miss the opportunities to build complex expressions explicitly from high-impact substructures. Identifying and reusing these could accelerate discovery by enabling recombination of partial solutions into globally correct forms.

A structural mechanism of this kind has a second advantage: because it scores substructures by their effect on fitness rather than by a target-specific model, it applies wherever a fitness function can be defined. Notably, this includes problems with no target variable at all. This matters because most symbolic regression systems specialize in either explicit mappings or implicit relations, but not both. Eureqa \citep{schmidt2009distilling}, the first dual-support approach, has remained closed sourced and now is integrated into a commercialized platform. We offer dual-support for both explicit and implicit symbolic regression where most modern models do not.

We introduce \textsc{SymMatika}, an evolutionary engine for symbolic discovery that combines feedback-based scheduling of evolutionary operators and the candidate generation distribution with an evolving library of high-impact symbolic substructures. Our contributions are summarized below:

\begin{itemize}
    \item An updating ``motif'' library that identifies and reuses high-impact symbolic substructures by quantifying their contribution to expression fitness through controlled ablation.
    \item Feedback-driven evolution where each population adapts both this operator mixture (i.e. crossover/mutation rates scheduled by fitness progress and plateau length) and its candidate-generation distribution (i.e. node-level probabilities updated from token frequencies in top-performing expressions), without any trained model.
    \item Dual-support for both explicit ($y = f(\mathbf{x})$) and implicit ($F(\mathbf{x}) = 0$) symbolic regression.
\end{itemize}

The structure of this paper is as follows: Sec.~\ref{sec:related} discusses related studies in symbolic regression, Sec.~\ref{sec:methods} describes our approach, and Sec.~\ref{sec:experiments} details experimental results as well as ablations and a detailed case study to quantitatively and qualitatively demonstrate the effectiveness of our ``motif'' library. Our full framework is open-sourced to support broader research in symbolic discovery.

\section{Related Works}
\label{sec:related}

\paragraph{Genetic Programming} GP \citep{koza1990genetic, koza1994genetic} evolves expression trees via biologically-inspired operations such as mutation and crossover. Operon \citep{burlacu2020operon} uses a steady-state GP model with tournament selection, while PySR \citep{cranmer2023interpretable} employs a multi-population island model and simulated annealing to balance exploration and exploitation during evolution; GP-GOMEA \citep{virgolin2021improving} and SBP-GP \citep{pawlak2014semantic} follow similar designs. These frameworks rely on fixed operator probabilities, which can hinder adaptive search in complex landscapes. \cite{guan2024feedback} use a modifiable crossover distribution to optimize crossover points, however it does not control the probability of crossover being selected over other genetic operations (i.e. mutation or single‑node crossover). \cite{doerr2019self} update mutation rates based on fitness‑success of offspring, yet it does not consider the overall evolutionary progress of the algorithm. \textsc{SymMatika} instead schedules the crossover/mutation mixture per island as a function of fitness progress and plateau length (Sec.~\ref{sec:feedback_gp}).

\paragraph{Symbolic Structural Reuse.}

A second family biases search with learned models to identify structural patterns in top-performing candidates. AI Feynman \citep{udrescu2020ai} uses neural networks to detect symmetries, units, and separability in the data, enabling recursive decomposition of complex equations. DSR \citep{petersen2019deep} instead trains an RNN policy to learn trends in expression coefficients. Neural-Guided Population Seeding (NGPS) \cite{mundhenk2021symbolic} seeds GP populations from that policy and update it from the best evolved expressions; uDSR \citep{landajuela2022unified} unifies these two strategies. Our approach instead adapts its candidate generation distribution without a trained model: we measure how often each operator appears in the top-performing expressions and raise its generation probability proportionally; e.g. if $\sin(\cdot)$ dominates the best candidates, new generated candidates are more likely to contain $\sin(\cdot)$ (Sec.~\ref{sec:motif}).

Capturing trends in top-performing candidates is useful not only at the token level but at the level of structure. Automatically Defined Functions (ADFs) \citep{koza1994genetic, ferreira2006automatically, zhong2015self} let each individual evolve reusable sub-functions alongside the program that calls them. However, incorporating ADFs can exacerbate the bloat problem \citep{turner2014cartesian}, where programs incrementally expand over evolutionary cycles. Modular GP-GOMEA \citep{harrison2505thinking} is a modern realization of this idea, where subexpressions are jointly evolved and reused as primitives. Capturing the modularity for symbolic regression (CaMo) \citep{liu2025camo} addresses the bloat problem and moves reuse to the population level, extending RNN-guided generation \citep{petersen2019deep, mundhenk2021symbolic} with an updating library of high-frequency modules sampled from top candidates, which become the new primitive set. However, frequency does not measure the substructure's contribution to fitness. \textsc{SymMatika} selects based on \textit{impact score}, measuring fitness differences from controlled ablations of subtrees from their parents, and maintains a fixed-size library of highest-impact subtrees. Because the library is bounded, the reuse mechanism itself does not grow over evolutionary cycles.

\paragraph{Symbolic Regression with Implicit Equations.} Most symbolic regression methods assume a labeled target and recover explicit mappings. Discovering implicit relations, such as conservation laws and other invariants, is comparatively underexplored. Eureqa \citep{schmidt2009distilling} introduces an implicit-derivative fitness that comparing symbolic and finite-difference partial derivatives, but it is closed-sourced nature and has low performance on modern benchmarks. Like Eureqa, Huynh et al. \citep{huynh2016semantics} support discovery of both explicit and implicit expressions, although it offers limited experimental results across 10 explicit and 3 implicit problems and no public implementation. PIE \citep{yufei2025advancing} maps point-set data to symbolic skeletons with learned priors from pre-training on explicit regression questions, while \textsc{SymMatika} requires no priors or supervised pre-training, and can immediately symbolically regress on unsupervised data. \textsc{SymMatika} adopts Eureqa’s implicit-derivative loss, and extend it with additional GP techniques. We compare against Eureqa on its published physical systems in Sec.~\ref{sec::eureqa_b}.

\section{Methods}
\label{sec:methods}

\textsc{SymMatika} is composed of two core components: (1) a multi-population, feedback-driven genetic programming engine, and (2) an updating library of high-impact symbolic motifs that capture reusable substructures. In this section, we introduce the formal setup and describe each component, followed by their integration for discovering both explicit and implicit expressions in data.

\subsection{Problem Setup}
\label{sec:problem_setup}

We represent symbolic expressions as algebraic trees $\tau$, with internal nodes as unary or binary operators (e.g., $+$, $\times$, $\cos$, $\log$) and leaf nodes as constants or variables. A pre-order traversal of $\tau$ yields the symbolic expression $f$, whose quality is assessed using task-specific fitness metrics.

\paragraph{Explicit relations.} Given data $\mathcal{D}_E = \{(\mathbf{x}_i, y_i)\}_{i=1}^N$ with $\mathbf{x}_i \in \mathbb{R}^d$, the goal is to find $f$ such that $y \approx f(\mathbf{x})$. Fitness can take the form of any supervised regression loss function (e.g. Mean Squared Error, Mean Log Error).


\paragraph{Implicit relations.} When target variables are not explicitly labeled, as is common in physical systems governed by invariants, we aim to discover expressions $f(\mathbf{x}) = 0$ that characterize underlying constraints or symmetries. Given a candidate expression \( f(x_1, x_2, \dots, x_n) = 0 \), to quantify fitness, we apply the implicit function theorem. Treating $x_i$ as an implicit function of $x_j$, the derivative is given by:
\[
\frac{\partial x_i}{\partial x_j} = -\frac{\frac{\partial f}{\partial x_j}}{\frac{\partial f}{\partial x_i}}
\]

We compute symbolic partial derivatives across all variable pairs and compare them against finite-difference numerical estimates. To account for variable interdependencies, which frequently occur in coupled dynamical systems, we generalize the paired partial derivative as:
\[
\frac{\partial x_i}{\partial x_j} = \frac{\partial x_i + \partial x_p \cdot \frac{\Delta x_p}{\Delta x_i}}{\partial x_j + \partial x_q \cdot \frac{\Delta x_q}{\Delta x_j}}
\]
where $x_p$ and $x_q$ are variables interdependent with $x_i$ and $x_j$, respectively. $\Delta$ represents finite differences in numerical analysis (including time‑series data), or the numerical approximation of the derivative, and $\partial$ represents the calculus partial derivative of a symbolic expression. We evaluate all possible such pairings and take the worst-case pairing for evaluation to penalize expressions that perform well only under selective dependencies. This approach targets alignment between the differential relationships implied by a candidate expression and those measured in the data, and an expression that reproduces how the variables co-vary captures the underlying constraint. \textsc{SymMatika} also uses numerical derivatives to safeguard against trivial and erroneous equations, where constant equations (i.e. equations with 0-value derivatives) and equations with near-zero or infinite derivatives are explicitly rejected.

Let $M_s$ and $M_n$ denote the symbolic and numerical paired-partial derivative matrices (shape $\binom{d}{2} \times N$). The implicit fitness is:
\[
\ell_{\mathcal{D}_I}(f) = -\frac{1}{N} \sum_{i=1}^{N} \log\left(1 + \left\| M_n(\mathbf{x}_i) - M_s(\mathbf{x}_i) \right\| \right)
\]

\subsection{Feedback-Based Genetic programming}\label{sec:feedback_gp}

Tree-based genetic programming uses two core operators: \textit{crossover}, which swaps subtrees at selected nodes between parents, and \textit{mutation}, which perturbs nodes within an individual. These are typically coupled with a selection mechanism (e.g. tournament selection) that favors high-fitness candidates. While effective, traditional GP applies static operator rates and rigid selection rules, often overlooking valuable substructures in lower-fitness individuals and failing to adapt to population dynamics. To address these limitations, we extend GP with three key mechanisms:

\paragraph{Single-node crossover.} Traditional subtree crossover introduces large structural changes. To enable finer control during late-stage optimization, we introduce single-node crossover, which swaps only one same-type node between trees (i.e. binary operation $\leftrightarrow$ binary operation, unary operation $\leftrightarrow$ unary operation, variable $\leftrightarrow$ variable, etc.). A constant $c$ is refined through scaling by $s \sim \text{U}(0.5,1.5)$ and then rounding to one decimal place. This keeps constants compact and interpretable at the cost of one-decimal precision (see App.~\ref{lim}).

\paragraph{Feedback-based operator scheduling.} Traditional GP systems use fixed rates for genetic operators such as mutation and crossover \citep{schmidt2009distilling}. While simple, static rates are suboptimal: coarse-grained changes like subtree crossover are useful early in evolution, whereas fine-grained adjustments like coefficient tuning are better suited for later stages. To enable adaptive behavior, we introduce feedback-based scheduling that dynamically adjusts operator probabilities based on the evolutionary context of each population.

Let $g_0$ be the average fitness of the top-$p$ individuals in a population prior to the GP loop, and let $g_n$ be the same statistic after $n$ generations. Let $h$ denote the number of consecutive generations with negligible fitness improvement (plateau), defined as $|g_{n} - g_{n-1}| < \epsilon$ for some threshold $\epsilon=1e^{-6}$. We define the operator probability function:
\[
    \mathbb{P}(g_n, h) = \begin{cases}
    m_i \pm (\frac{|g_0 - g_n|}{g_0} \cdot|m_f - m_i|) \pm 0.1h & h \ge 2 \\
    m_i \pm (\frac{|g_0 - g_n|}{g_0} \cdot|m_f - m_i|) & \text{else}
\end{cases}
\]
Here, $m_i$ and $m_f$ are the initial and final operator probabilities for each genetic operation (e.g. crossover, mutation), with signs determined by whether each operator frequency should increase or decrease over time. Crossover starts with probability 60\% and decreases to 5\%, single-node crossover starts with probability 10\% and increases to 15\%, mutation starts with probability 30\% and increases to 80\%. This function is safe-guarded by these probability bounds and our probabilistic selection model, so it is resilient to abnormal degradations in average fitness.

\paragraph{Island populations with migration.} We evolve a set of populations or ``islands'' in parallel. The island model for GP is a powerful tool for simultaneously evolving multiple distinct evolutionary paths \citep{duarte2017dynamic, whitley1999island, cranmer2023interpretable}. We allow for small migrations between islands by swapping small subsets of individuals between populations. Initially, 1\% of each population is swapped every 20 iterations. These subsets increase to 2\% according to a growing control rate $\alpha_M$. We keep migrations small and infrequent to promote population diversity and prevent convergent evolution, which we tend to observe when migration rates exceed 3\%.

\subsection{Symbolic Substructure Reuse}\label{sec:motif}

After each iteration of GP, we generate new candidates to replace subsets of worst-performing individuals in each population. Candidate generation in \textsc{SymMatika} is modeled as a parameterized distribution over expression trees, denoted $p(\tau \mid \theta)$, where $\tau$ is a symbolic expression and $\theta$ encodes node-level generation probabilities (e.g., for selecting constants, variables, and operators) \citep{mundhenk2021symbolic, petersen2019deep}. These parameters are updated based on frequency feedback from high-fitness individuals: for each operator type $\theta_i$, we compute its frequency $f_{\theta_i}$ across top-performing expressions, and apply the update:
\[
\theta_i \leftarrow \theta_i + \beta_{\theta_i} f_{\theta_i}
\]
Here, $\beta_{\theta_i}$ is a scaling factor that accounts for natural biases, e.g., binary operators ($\beta_{\theta_i}=0.2$) such as $+$ and $\times$ may occur more often than unary functions ($\beta_{\theta_i}=1.0$) like $\log$ or $\exp$, and are thus adjusted more conservatively. These biases were selected from observations and tuning of our parameterized candidate generator during initial testing. This feedback mechanism enables population-specific adaptation of the generative distribution, which can bias expression construction toward promising token-level patterns.

However, frequency-based feedback alone is limited: it captures which tokens are useful, but cannot be used to explicitly discover structural patterns or partial solutions in individuals. For instance, in a symbolic regression problem with correct solution $f =x^2 + y^2 + z^2$, a partial solution $g = x^2 + y^2 - \cos(z)$ would not be recognized as having some correct terms through frequency analysis. To better identify and reuse partial solutions, we introduce a substructure discovery mechanism inspired by biological sequence ``motifs.''

In molecular biology, a sequence ``motif'' is a recurring nucleotide or amino-acid pattern with functional significance \citep{tateno1997evolutionary, liu2002statistical, chou2011biological, grant2021xstreme}. Analogously, we define a symbolic \textit{motif} as a high-impact subexpression, i.e. a subtree within an expression that contributes strongly to its overall fitness. We use this intuition to construct a reusable library of symbolic motifs and employ them for guided recombination.

\begin{figure}[t]
    \centering
    \includegraphics[width=\linewidth]{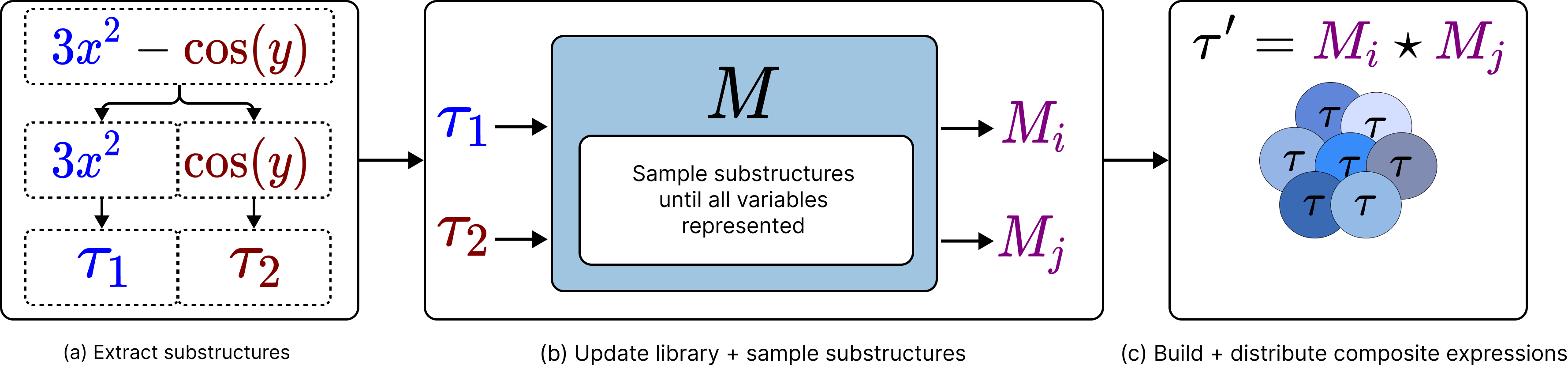}
    \caption{Motif-based recombination across evolving populations with shared top performers.}
    \label{fig:motif-library}
    \vspace{-15pt}
\end{figure}

\paragraph{Initialization and generation.} To implement structural reuse, we maintain a central library $M$ that stores high-impact subexpressions across populations; $M$ is an impact-ordered list with $k_M$ entries, typically $k_M \le 50$. Each entry in the library $M_i$ consists of a pair $(\tau_i, I(\tau_i))$, where $0 \le i \le k_M$. Here, $\tau_i$ is a subtree and $I(\tau_i)$ is its estimated impact score on fitness. We calculate impact score with:

\begin{equation}
\label{eq:impact}
    I(\tau_i) = \mathcal{L}(f - \tau_i) - \mathcal{L}(f)
\end{equation}

where $\mathcal{L}(f)$ is the loss of the original expression $f$, and $\mathcal{L}(f - \tau_i)$ is the loss of $f$ ablating $\tau_i$ --- note: we replace $\tau_i$ with an ``identity'' (e.g. $0$ for $+$, $1$ for $\times$). $I(\tau_i)$ quantifies the importance of the substructure to the overall quality of the expression. Furthermore, for both explicit and implicit relations (Sec.~\ref{sec:problem_setup}), we take absolute values of each term in ~\ref{eq:impact}, so the calculation is consistent for both fitness functions. After each generation, we examine the top-$p$ individuals from each island population to identify high-impact subtrees. For a given expression $\tau$ (e.g., $\tau = 3x^2 -\cos(y)$), we extract the left and right subtrees of each internal node (e.g., $\tau_L = 3x^2$, $\tau_R = -\cos(y)$. Note: $\cos(y)$ is a negative term in the example, so we extract it as $-\cos(y)$) and assess their fitness contribution. Newly extracted ``motifs'' are compared to existing entries, high-impact subtrees replace low-impact ones, and the library is re-sorted by impact score.

\paragraph{Composite synthesis and co-evolution.} Once the library is updated, composite expressions are synthesized by sampling ``motifs'' until all system variables are represented. These are composed into full expressions using randomly selected binary algebraic operators $+$, $-$, $\times$, or $\div$. The resulting expressions are inserted into a dedicated ``motif'' population $P_M$, which co-evolves alongside the main island populations. To propagate promising structural components throughout the evolutionary process, high-performing expressions from $P_M$ are periodically injected back into the main populations. This allows strong partial solutions, discovered independently across islands and generations, to be recombined and reused in novel ways, ultimately improving both convergence speed and expression quality.

This structure-aware mechanism complements token-level parameter updates by identifying semantically meaningful subtrees for reuse and recombination. It accelerates symbolic discovery by enabling partial solution reuse and coordinated variable coverage. Motif recombination also improves robustness to local optima by synthesizing expressions from independently validated high-impact components. We visualize this pipeline in Fig.~\ref{fig:motif-library}, outline the full algorithm in Alg.~\ref{algorithm}, visualize the full model in Fig.~\ref{fig:main-model}, and list hyperparameters in Appendix Tab.~\ref{hyperparameters}.

\vspace{-5pt}

\subsection{Implementation Details}
\label{sec::implementation}

\textsc{SymMatika} is a multi-core \texttt{C++23} library, compiled with \texttt{clang++} and optimized using \texttt{-O3} and \texttt{-march=native} flags. It leverages \texttt{SymEngine} for symbolic computation, \texttt{Eigen} for linear algebra and paired-partial derivative operations, and \texttt{OpenMP} for parallelization. The full framework is open-sourced to support future work in interpretable modeling and symbolic discovery. All experiments (Sec.~\ref{sec:experiments}) are run with random seeds and hyperparameters specified in Sec.~\ref{hyperparameters} on a 2023 Apple M3 Max MacBook Pro with a 14-core CPU, 30-core GPU, and 36GB unified memory.

\section{Experiments}
\label{sec:experiments}

In this section, we present our experimental results that evaluate the effectiveness and simplicity of \textsc{SymMatika} in both explicit and implicit symbolic regression. We isolate the contribution of each component with detailed ablations, and present a case study examining what the motif library learns and where reuse helps. We detail our hyperparameter set in Tab.~\ref{hyperparameters} and we list the full supported primitive/function set supported by \textsc{SymMatika} in App.~\ref{primitives}.


\subsubsection{Datasets}

We evaluated on four benchmark families: the Nguyen suite \citep{uy2011semantically}, the 119 Feynman equations \citep{udrescu2020ai}, the SRBench Black-box problems, and the four physical systems released with Eureqa \citep{schmidt2009distilling} for implicit discovery. We adhere to the evolutionary search budget defined by \cite{petersen2019deep} for Nguyen of 2M expression evaluations, and the SRBench protocol for Feynman and Black-box; since the Eureqa problem set has no standardized budget, we adhere to Nguyen's. To assess how methods leverage textual priors, recent benchmarks like LLM-SRBench \citep{shojaee2025llm} pair tasks with natural-language scientific context. However, because \textsc{SymMatika} operates strictly on observational data, our primary baseline is \cite{shojaee2025llm}'s data-only benchmark standard (PySR), which we evaluated directly in our experiments.

\subsubsection{Evaluation Metrics}

We used three evaluation metrics:

\begin{itemize}
    \item \textbf{Exact recovery rate} for Nguyen and Eureqa (i.e. exact symbolic equivalence \citep{mundhenk2021symbolic}); we also reported test $R^2$ and maximum complexity on Nguyen for GP-baselines we ran ourselves.
    \item \textbf{Error \& Complexity}, following SRBench \citep{la2021contemporary}, we plotted median $R^2$ / mean $R^2>0.99$ proportion vs. equation complexity (i.e. number of nodes in an expression tree, or total number of variables, constants, and operators) for 119 Feynman and 57 Black-box problems.
    \item \textbf{Wall-clock runtime} for the Nguyen and Eureqa problems.
\end{itemize}

All experimental results were reported over independent random seeds with 95\% normal-approximation (Wald) confidence intervals (CIs), and we perform statistical testing on Feynman and Black-box with Wilcoxon signed-rank test ($\alpha=0.05$) \citep{la2021contemporary} where a baseline's performance is close to ours. Furthermore, we conducted detailed ablation studies isolating each component's impact on model performance in Sec.~\ref{sec:motif_analysis}.


\subsection{Nguyen Benchmark}\label{sec::nguyen_b}

\paragraph{Exact recovery.} Tab.~\ref{tab:nguyen} reports exact recovery rates over 100 independent runs per problem. We took reported results from NGPS and DSR and ran our own evaluations of PySR, Operon, GP-GOMEA, and SBP-GP with consistent experimental settings from \cite{petersen2019deep}. Because results for NGPS, DSR, and Eureqa are available only as reported per-problem rates, and not as matched runs, we compare methods without reporting significance tests on this benchmark. We report an average recovery rate of \textbf{96.5\%}, outperforming all other methods, and \textbf{61\%} recovery rate on Nguyen-12. Notably NGPS \citep{mundhenk2021symbolic} report a 12\% success rate on a relaxed variant with expanded data range (Nguyen-$12^\star$), although our performance remains significantly stronger on the original Nguyen-12 dataset. These results highlight the advantage of combining structural motif reuse with feedback-driven parameter tuning, especially for long and composite expressions; we demonstrate this in Sec.~\ref{sec:motif_analysis}.

\paragraph{Fit, Complexity, \& Runtime}
\label{sec:nguyen_runtime}

Tab.~\ref{tab:nguyen_runtime} reports test $R^2$, wall-clock time to reach the fit threshold ($R^2>0.999$), and expression complexity over 10 trials per problem for \textsc{SymMatika} and the GP-based methods we ran under identical settings, stopping rule, and hardware (PySR, Operon, GP-GOMEA, SBP-GP), with medians and maxima across trials. \textsc{SymMatika} returned the most compact expressions (average complexity 10.1, against 13.7 for PySR and 21.2 for GP-GOMEA) with a median fit time of 3.14 sec. Operon was faster at 1.3 sec, but returned expressions twice as complex. We also list the values reported for CaMo \citep{liu2025camo}, which has no public implementation. These use 100 samples per problem against our 20 on different hardware, so the comparison is between two full systems and does not isolate the effect of impact-based selection, which Sec.~\ref{sec:motif_analysis} and Tab.~\ref{tab:ablations} test directly. Tab.~\ref{tab:nguyen-runtime} separates time to fit from time to exact recovery, and Tab.~\ref{tab:nguyen-per-run} lists per-run expressions for Nguyen-7 and Nguyen-12.

\subsection{Feynman}
\label{sec:feynman}

The Feynman equations are sampled to form datasets $\mathcal{D} = \{(\mathbf{x}_i, y_i)\}_{i=1}^N$. We recorded the error-complexity plot (mean $R^2>0.99$ proportion vs. equation complexity) against SRBench \citep{la2021contemporary} results.

\begin{figure}[t]
\centering
\includegraphics[width=\linewidth]{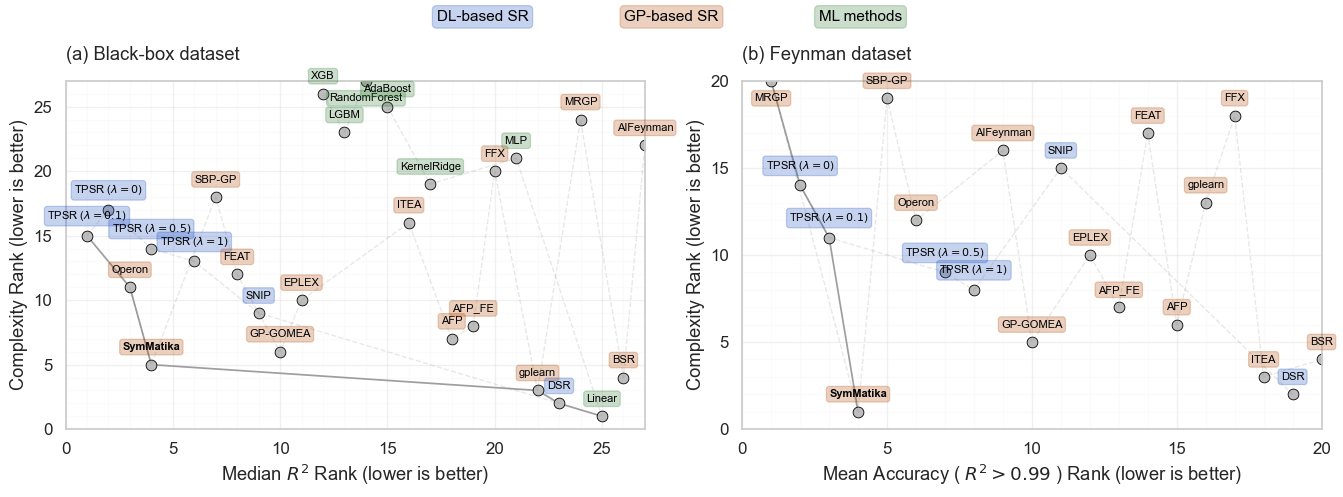}
\caption{Error–complexity (rank) trade-off on the Black-box (left) and Feynman (right) datasets.}
\label{fig:errorcomplexity}
\end{figure}

Across 5 trials, \textsc{SymMatika} achieves a mean $R^2>0.99$ proportion of 0.942 with a mean complexity of 13.294 (Fig.~\ref{fig:errorcomplexity}). Only MRGP \citep{arnaldo2014multiple} is more accurate (0.977), but its expressions are about $276\times$ more complex (3669.357). Against the remaining SRBench methods, \textsc{SymMatika}'s accuracy is significantly higher (two-tailed Wilcoxon signed-rank test, $p<0.05$), except from Operon (0.940) and SBP-GP (0.941), in which it is slightly improved ($p \sim 0.9$). Notably, those two methods return expressions about $5\times$ (69.875) and $37\times$ (489.416) more complex. On the Black-box subset, however, we reported much better results in both error and complexity than MRGP. Additionally, two versions of TPSR \citep{shojaee2023transformer} slightly outperformed our mean ($R^2>0.99$) accuracy with 0.949 (TPSR[$\lambda=0.1$]) and 0.952 (TPSR[$\lambda=0$]) to our 0.942, however our complexity was significantly improved with our 13.294 compared to their 84.42 (TPSR[$\lambda=0$]) and 57.22 (TPSR[$\lambda=0.1$]). Additionally, we outperformed SNIP \citep{meidani2023snip} at mean $R^2>0.99$ proportion 0.882 and complexity 31.630. \textsc{SymMatika} Pareto dominated other symbolic regression algorithms (e.g. SNIP, TPSR [$\lambda=0.5, \lambda=1$]) and was not dominated by any other algorithm \citep{fongpareto}; additionally, we led the front in model complexity. We reported all algorithm performances with 95\% confidence intervals in Fig.~\ref{fig:feynman_bb} and Tab.~\ref{tab:sidebyside}.

\subsection{Black-box}

We evaluated on 57 Black-box problems gathered from SRBench used by TPSR \citep{shojaee2023transformer} which include synthetic (e.g. synthetic pollen dataset) and real-world (e.g. historical harvest yields in Lake Erie vineyard) datasets and we recorded median $R^2$ and mean equation complexity for the error-complexity plot (Fig.~\ref{fig:errorcomplexity}). We ran 5 full trials and recorded median $R^2$ of 0.931 and equation complexity of 28.059. We outperformed nearly all benchmarked algorithms in median $R^2$, including SNIP at 0.853 (complexity 47.527), and matched TPSR [$\lambda=0.5$] (complexity 82.58), where $\lambda$ is a hyperparameter balancing fitting and complexity reward), and falling slightly behind TPSR [$\lambda=0$] at 0.938 (complexity 129.85), Operon at 0.937 (complexity 61.74) and TPSR [$\lambda=0.1$] at 0.945 (complexity 95.71). Our complexity significantly improved on these algorithms by a factor of \textbf{3-4.5} for TPSR and a factor of \textbf{2} for Operon, and MRGP which reports a much-lower median $R^2$ of 0.502 larger and complexity 9878.172. TPSR records one trial so we cannot report meaningful $p$-testing given limited experimental data, yet we could report that the Operon’s outperformance of us is not statistically significant, i.e. two-tailed $p=0.241$ with Wilcoxon signed-rank test. \textsc{SymMatika} Pareto-dominated many algorithms, and placed strongly on the Pareto-front; notably certain models (e.g. DSR, gplearn) are very simple and so cannot be dominated, yet their accuracy was extremely low.

\subsection{Eureqa Physical Systems}\label{sec::eureqa_b}

We evaluated implicit discovery on the four time-series physical systems released with Eureqa \citep{schmidt2009distilling} with the same experimental setup as Nguyen. Unlike the explicit benchmarks (i.e. Nguyen, Feynman \& Black-box), these provide no target variable: recovery involves discovering hidden algebraic constraints $F(\mathbf{x})=0$, $\mathbf{x} \in \mathbb{R}^n$ from raw trajectory data. Notably, we acknowledged that Eureqa's specific implementation and hardware for experiments are unknown, and so we make no direct runtime comparisons between algorithms. In Tab.~\ref{eureqa}, we reported exact symbolic recovery rates (i.e. Success (\%)), mean MLE, and mean wall-clock runtime to exact recovery on each problem; note that failure cases (i.e. exact recovery not reached) were ran until the evolutionary search budget was expended (see Tab.~\ref{hyperparameters} for more details. We report the exact symbolic targets and some recovered expressions for each problem in Tab.~\ref{tab:eureqa_targets}. With this, we demonstrate \textsc{SymMatika}'s capability to support both explicit and implicit symbolic regression.

\begin{table}[t]
\centering
\caption{Performance on the Eureqa implicit regression problems}
\label{eureqa}
\small
\begin{tabular}{l r r r}
\toprule
Problem & Success (\%) & MLE mean $\pm$ sd & Runtime mean $\pm$ sd (s) \\
\midrule
\texttt{circle\_1} & $100$ & $-2.15 \times 10^{-13} \pm -1.6\times10^{-16}$ & $1.14 \pm 0.02$ \\
\texttt{pendulum\_h\_1} & $90$ & $-0.0142 \pm 0.007$ & $3.01 \pm 1.41$\\
\texttt{sphere\_1} & $100$ & $-0.0225 \pm 0.0$ & $14.62 \pm 1.83$ \\
\texttt{dbl\_pendulum\_h\_1} & $10$ & $-0.1795 \pm 0.0734$ & $905.23 \pm 17.76$ \\
\bottomrule
\end{tabular}
\end{table}

\subsection{Analysis of the Motif Library: Ablations \& Case Study}
\label{sec:motif_analysis}

The ``motif'' library is \textsc{SymMatika}'s central evolutionary contribution, so we evaluated it in two ways: (1) ablations on Nguyen and Feynman problems to measure its effect on recovery, and (2) an inspection of the library's contents to showcase what structures are captured.

\paragraph{Nguyen ablation.} We evaluated eight configurations on four hard Nguyen problems (i.e. 4, 5, 7, 12): \textbf{(1)} a baseline with standard GP (no parameter tuning or motif reuse), \textbf{(2)} baseline with adaptive operator rates, \textbf{(3)} baseline with shared-island library, \textbf{(4)} baseline with $\theta$ -generation, \textbf{(5-7)} baseline with ``motif'' library (random subtree selection, frequency-based scoring, and impact-based scoring), and \textbf{(8)} the full model with both mechanisms enabled. Each was run with 50 independent seeds and results were reported in Tab.~\ref{tab:ablations}. \textbf{(8)} consistently achieved high recovery rates, and beat all other configurations with $p<0.05$. \textbf{(4)} provided consistent boosts across all benchmarks, while motif reuse was especially helpful on longer expressions (e.g. Nguyen-12). We observed that \textbf{(7)}, impact-based scoring, reported a higher average than frequency-based and random subtree selection with $p<0.05$, demonstrating the strength of impact-based scoring over frequency \citep{liu2025camo}. Our results confirmed that token-level trends and reusable substructures capture complementary inductive biases that are critical for recovering long or irregular expressions. We discuss limitations in depth in Appendix Sec.~\ref{lim}, and report the extra computational cost of scoring ``motifs'' in Tab.~\ref{tab:impact-cost}.

\begin{table}[t]
  \centering
  \small
  \setlength{\tabcolsep}{3pt}
  \caption{Nguyen recovery rates (\%) over 50 independent trials per problem. \emph{Left:} component ablation. \emph{Right:} motif-selection ablation, i.e. impact-scored, frequency-scored, \& random subtree selection}
  \label{tab:ablations}
  \begin{tabular}{@{}l rrrr rrr r@{}}
    \toprule
    & \multicolumn{4}{c}{Component ablation} & \multicolumn{3}{c}{Motif selection rule} & \\
    \cmidrule(lr){2-5} \cmidrule(l){6-8}
    \textbf{Problem} & \textbf{Baseline} & \textbf{Adaptive} & \textbf{Shared} & \textbf{$\theta$-Gen.} & \textbf{Random} & \textbf{Frequency} & \textbf{Impact} & \textbf{Full Model} \\
    \midrule
    Nguyen-4  & 44 & 50 & 28 & 48 & 44 & 34 & 74 & 98 \\
    Nguyen-5  & 100 & 100 & 92 & 92 & 88 & 100 & 100 & 100 \\
    Nguyen-7  & 72 & 76 & 66 & 94 & 64 & 52 & 80 & 94 \\
    Nguyen-12 & 6 & 6 & 10 & 2 & 18 & 16 & 34 & 42 \\
    \midrule
    \textbf{Average} & 55.5 & 58.0 & 49.0 & 59.0 & 53.5 & 50.5 & \textbf{72.0} & \textbf{83.5} \\
    \bottomrule
  \end{tabular}
  \vspace{-10pt}
\end{table}

\paragraph{Feynman ablation.} To evaluate beyond Nguyen, we compared between baselines \textbf{(4)} (baseline + $\theta$-based generation) and \textbf{(8)} (full model) on three Feynman equations with clear compositional structure, i.e. I.12.11 ($F = q(E_f + Bv\sin\theta)$), I.13.4 ($K = \tfrac{1}{2}m(v^2 + u^2 + w^2)$), and II.11.20 ($P_* = \frac{n_\rho p_d^2 E_f}{3 k_b T}$), over 20 seeds each. Without the library, \textsc{SymMatika} recovered I.12.11 in 13/20 runs, I.13.4 in 15/20, and II.11.20 in 8/20; with it, 20/20 on both I.12.11 and I.13.4, with 18/20 on II.11.20.

\begin{table}[H]
\centering
\caption{Highest-impact motif library entries for three Feynman equations, from a single run of each. $I(\tau)$ is the fitness change from ablating the subtree from its parent expression; entries are listed in decreasing impact.}
\label{tab:motifs}
\begin{tabular}{@{}llr@{}}
\toprule
Equation (target) & Motif $\tau$ & $I(\tau)$ \\
\midrule
\multirow{3}{*}{I.12.11 \; $F = q(E_f + Bv\sin\theta)$}
  & $B - v + B \cdot q \cdot v \cdot \sin(\theta)$ & 45.7 \\
  & $-0.9 \cdot B \cdot q \cdot v \cdot \sin(B - (B + \theta))$ & 17.1 \\
  & $q \cdot E_f$ & 16.6 \\
\midrule
\multirow{3}{*}{I.13.4 \; $K = \tfrac{1}{2}m(v^2+u^2+w^2)$}
  & $v^2 + u^2 + w^2$ & 201.2 \\
  & $0.5 \cdot m \cdot v^2 + 0.5 \cdot m \cdot w^2$ & 175.8 \\
  & $0.5 \cdot w^2$ & 24.1 \\
\midrule
\multirow{3}{*}{II.11.20 \; $P_* = \frac{n_\rho p_d^2 E_f}{3 k_b T}$}
  & $0.33 \cdot \frac{E_f \cdot p_d^2}{T \cdot k_b}$ & 4676.0 \\
  & $0.3 \cdot \frac{E_f}{k_b}$ & 347.1 \\
  & $0.3 \cdot E_f \cdot p_d^2 \cdot n_\rho$ & 42.0 \\
\bottomrule
\end{tabular}
\end{table}

\paragraph{What the library learns.} Tab.~\ref{tab:motifs} lists the three highest-impact library entries for I.12.11, I.13.4, and II.11.20, each from a single run, with impact scores $I(\tau)$. In each case, the library isolated the target's main components. For I.12.11 the two additive terms of the target, $Bqv\sin\theta$ and $qE_f$ were identified. For I.13.4 the top entry was exactly $v^2 + u^2 + w^2$, and the second was a partial sum $\tfrac{1}{2}m(v^2 + w^2)$. For II.11.20 the top entry was the full target with one factor missing. Across all three runs, we demonstrated qualitatively how the ``motif'' library learns and frequently reuses meaningful physical sub-structures.

\section{Conclusions}

We introduce \textsc{SymMatika}, a symbolic regression framework that combines feedback-guided genetic programming with an evolving library of high-impact symbolic substructures to recover both explicit and implicit expressions from data with the same search mechanisms. Across benchmarks, \textsc{SymMatika} recovered 96.5\% of Nguyen expressions and 61\% of runs on Nguyen-12, where no other method exceeded 2\%, it lies on the error–complexity Pareto front for the Feynman and Black-box problems with the most compact expressions of any competitive method, and we demonstrated implicit relations discovery on Eureqa's published physical systems data. We highlighted the strengths of high-impact substructure reuse with controlled ablations, and, through inspection of the library, we showed that it extracts meaningful subexpressions of the full symbolic target. The consistently low complexity of returned expressions matters, as compact expressions are the interpretable ones, and this underlines the point of symbolic regression for science. Future work will explore discovering structural trends in long expressions with few reusable substructures, as well as more testing of complex floating-point constants.

\subsubsection*{Broader Impact Statement}
\textsc{SymMatika} can enable scientists to model complicated physical systems with compact, interpretable symbolic equations. Through experimental results and detailed ablation studies, we have demonstrated the strength of impact-based substructure reuse for symbolic discovery, and through a case-study (see Sec.~\ref{sec:motif_analysis}) we show that these discovered substructures are easily interpretable. This can help scientists discover specifically \textit{which} symbolic substructures contribute most to the model’s fitness.

\newpage

\bibliography{main}
\bibliographystyle{tmlr}

\appendix
\section{Appendix}

\subsection{Sample Outputs From Subset of Feynman Equations}\label{sample-output}

\begin{table}[H]
  \centering
  \caption{Comparison of ground-truth vs.\ discovered equations.}
  \label{tab:equations}
  \small
  \setlength{\tabcolsep}{4pt}
  \begin{tabular}{@{}l p{0.42\columnwidth} p{0.42\columnwidth}@{}}
    \toprule
    \textbf{Equation Number}
      & \textbf{Ground Truth Equation}
      & \textbf{Discovered Equation}\\
    \midrule
    I.18.4 & $r = \frac{m_1r_1+m_2r_2}{m_1+m_2}$ & $r = \frac{r_1 m_1}{\left(m_1 + m_2\right)} + \frac{r_2 m_2}{\left(m_1 + m_2\right)}$ \\
    II.11.20 & $P_*=\frac{0.333333333333333 E_f p_d^2 n_{\rho}}{T k_b}$ & $P_* = \frac{n_{\rho}p_d^2E_f}{3k_bT}$ \\
    I.37.4 & $I_* = I_1 + I_2 + 2\sqrt{I_1I_2}\cos{\left(\delta\right)}$ & $I_* = I_1 + I_2 + 2\sqrt{I_1I_2}\cos{\left(\delta\right)}$ \\
    I.24.6 & $E_n=\frac{1}{4}m(\omega^2+\omega_0^2)x^2$ & $E_n=\frac{1}{2}m(\omega^2+\omega_0^2)\frac{1}{2}x^2$ \\
    III.17.37 & $f=\beta(1+\alpha\cos(\theta))$ & $f=\beta + \beta \cos{\left(\theta\right)} \alpha$ \\
    I.27.6 & $f_f=\frac{d_2}{\left(n + \frac{d_2}{d_1}\right)}$ & $f_f=\frac{1}{\frac{1}{d_1}+\frac{n}{d_2}}$ \\
    I.47.23 & $c=\sqrt{\frac{\gamma pr}{\rho}}$ & $c=\sqrt{\frac{\gamma pr}{\rho}}$ \\
    I.12.11 & $F=q(E_f+Bv\sin(\theta))$ & $F=qE_f+qBv\sin(\theta)$ \\
    \bottomrule
  \end{tabular}
\end{table}

We demonstrate outputs on a subset of the Feynman Equations and compare ground truth equations to discovered equations. The ground truth equations and discovered equations are syntactially different although mathematically equivalent. These discoveries were found with implementation details consistent with all other experiments. All discovered equations record a Mean Log Error (MLE) loss of $\le10^{-16}$.

\subsection{\textsc{SymMatika} Algorithm}

\begin{algorithm}[H]
\caption{\textsc{SymMatika} Algorithm}
\label{algorithm}
\textbf{input:} Symbolic regression problem with data $\mathcal{D}$ and relation type $t$

\textbf{output:} Best-fitting expression $\tau^{\star}$
\begin{algorithmic}[1]

\STATE Initialize islands $I_1, \dots, I_m$ with populations $P_1, \dots, P_m$ and node parameters $\theta_1, \dots, \theta_m$
\STATE Initialize fitness trackers $F_1, \dots, F_m$
\STATE Define motif library $M$

\FOR{each generation $g$}
    \FOR{each island $I_i$}
        \STATE $E(I_i) \gets$ evolve island population $P_i$ with $k$ runs of GP
        \STATE $\tau_u \gets$ update island parameters $\theta_i$
        \IF{plateau height $h \geq h_{\text{max}}$}
            \STATE rebuild population $P_i$
        \ENDIF
        \STATE $\tau_m \gets$ update motif library $M$ with high-performing subexpressions
        \STATE $I_i \gets$ migrate individuals between islands with growing rate $\alpha_M$
    \ENDFOR
    \STATE $E(M) \gets$ evolve motif population and distribute fittest individuals to island populations
\ENDFOR
\STATE return $\tau^{\star}$
\end{algorithmic}
\end{algorithm}

\subsection{Tabulated Parameters}\label{hyperparameters}

\textsc{SymMatika} is initialized with population sizes of 10,000 individuals per island, followed by truncation to the top 400 for the GP-loop. The number of islands $I$ is set proportional to $I \propto 2d$, where $d$ is the number of variables (capped at $I = 8$ for large SR-problems with five or more variables due to computational considerations). We run each experiment for 1,500 generations of GP.

\begin{table}[H]
    \centering
    \caption{Hyperparameter values for all experiments.}
    \resizebox{\linewidth}{!}{%
    \begin{tabular}{@{}l c c l@{}}
        \toprule
        \textbf{Hyperparameter} & \textbf{Symbol} & \textbf{Value} & \textbf{Comment} \\
        \midrule
        \multicolumn{4}{@{}l}{\textbf{GP Parameters}} \\
        \midrule
        Initial population size & -- & 10,000 & see Sec.~\ref{sec:experiments} \\
        Truncated population size & -- & 400 & see Sec.~\ref{sec:experiments} \\
        Generations & $G$ & 1500 & -- \\
        GP evaluations per generation & $k$ & 200 & see Fig.~\ref{fig:main-model} \\
        Island populations & $m$ & $\max(2d,8)$ & $d$ is number of variables \\
        Migration rate & $\alpha_M$ & 0.02 & migration between islands (Algorithm~\ref{algorithm}) \\
        Problem type & $t$ & $\{0,1\}$ & explicit or implicit relation (Algorithm~\ref{algorithm}) \\
        Initial operator probabilities & $m_i$ & $(0.60,\ 0.10,\ 0.30)$ & crossover, single-node crossover, mutation \\
        Final operator probabilities & $m_f$ & $(0.05,\ 0.15,\ 0.80)$ & crossover, single-node crossover, mutation \\
        \midrule
        \multicolumn{4}{@{}l}{\textbf{Motif Library Parameters}} \\
        \midrule
        Library size & $k_M$ & 50 & maximum number of ``motifs'' in library \\
        Top-ranked candidates & $p$ & 10 & candidates used for motif library analysis \\
        Natural bias per node type & $\beta_{\theta}$ &
            $(\underbrace{0.2, \dots, 0.2}_{5\ \text{times}},\ \underbrace{1.0, \dots, 1.0}_{13\ \text{times}})$ &
            bias term for node type $\theta_i$ (Sec.~\ref{sec:motif}) \\
        \bottomrule
    \end{tabular}
    }
    \label{tab:settings}
\end{table}

We initialize \textsc{SymMatika} with these hyperparameters, and in experiments, we run until we hit the evolutionary search budget for the respective baseline (e.g. 2M expression evaluations for Nguyen and Eureqa).

\subsection{Primitive Set}
\label{primitives}

\textsc{SymMatika} builds expression trees over the primitive set in Tab.~\ref{tab:primitives}. Internal nodes are elements from the primitive set, and leaf nodes are problem variables (i.e. $x_1,\dots,x_n$) or constants. This operator set can be customized freely by the user.

\begin{table}[h]
\centering
\caption{Primitive operator set.}
\label{tab:primitives}
\begin{tabular}{lll}
\toprule
Node type & Default & Extended (opt-in) \\
\midrule
Binary operators & $+,\;-,\;\times,\;\div,\;\wedge$ & --- \\
Unary operators  & $\sin,\;\cos,\;\tan$ & $\sinh,\;\cosh,\;\tanh$ \\
                 & $\log_{10},\;\ln,\;\sqrt{\cdot},\;\exp$ & $\arcsin,\;\arccos,\;\arctan$ \\
Leaf        & variables $x_1,\dots,x_d$; constants & --- \\
\bottomrule
\end{tabular}
\end{table}

\subsection{Experimental Results}\label{appx_results}

\begin{table}[H]
\centering
\caption{Exact recovery rates (\%) on the Nguyen benchmark over 100 runs, with 95\% confidence intervals on the average. \textsc{SymMatika} achieves the highest average (96.5\%) and is the only method to recover Nguyen-12 at a substantial rate (61\%).}
\label{tab:nguyen}
\footnotesize
\setlength{\tabcolsep}{3pt}
\resizebox{\textwidth}{!}{%
\begin{tabular}{@{}l>{\scriptsize}l rrrrrrrr@{}}
\toprule
\textbf{Problem} & \textbf{Expression} & \textbf{SymMatika} & \textbf{NGPS} & \textbf{DSR} & \textbf{Eureqa} & \textbf{GP-GOMEA} & \textbf{SBP-GP} & \textbf{PySR} & \textbf{Operon} \\
\midrule
Nguyen-1  & $x^3 + x^2 + x$                    & 100 & 100 & 100 & 100 & 100 &  99 & 99 & 34 \\
Nguyen-2  & $x^4 + x^3 + x^2 + x$              & 100 & 100 & 100 & 100 & 100 & 100 & 99 & 28 \\
Nguyen-3  & $x^5 + \cdots + x$                 & 100 & 100 & 100 &  95 & 100 & 100 & 78 & 2 \\
Nguyen-4  & $x^6 + \cdots + x$                 & 100 & 100 & 100 &  70 & 100 & 100 & 34 & 1 \\
Nguyen-5  & $\sin(x^2)\cos(x) - 1$             & 100 & 100 &  72 &  73 &  92 &  95 & 27 & 3 \\
Nguyen-6  & $\sin(x) + \sin(x + x^2)$          & 100 & 100 & 100 & 100 & 100 & 100 & 99 & 25 \\
Nguyen-7  & $\log(x + 1) + \log(x^2 + 1)$      &  98 &  97 &  35 &  85 &  81 &  85 & 11 & 0 \\
Nguyen-8  & $\sqrt{x}$                         & 100 & 100 &  96 &   0 & 100 & 100 & 77 & 51 \\
Nguyen-9  & $\sin(x) + \sin(y^2)$              & 100 & 100 & 100 & 100 & 100 & 100 & 99 & 71 \\
Nguyen-10 & $2\sin(x)\cos(y)$                  & 100 & 100 & 100 &  64 & 100 & 100 & 89 & 58 \\
Nguyen-11 & $x^{y}$                            & 100 & 100 & 100 & 100 & 100 & 100 & 42 & 24 \\
Nguyen-12 & $x^4 - x^3 + \tfrac{1}{2}y^2 - y$  & \textbf{61} & 0 & 0 & 0 & 0 & 0 & 2 & 0 \\
\midrule
\textbf{Average} & & \textbf{96.5}\,$\pm$0.6 & 91.4\,$\pm$1.6 & 83.6\,$\pm$1.8 & 73.9\,$\pm$2.0 & 89.4\,$\pm$1.6 & 89.9\,$\pm$1.6 & 63.0\,$\pm$2.0 & 29.7\,$\pm$1.3 \\
\bottomrule
\end{tabular}%
}
\end{table}

\begin{table}[t]
\centering
\caption{Test $R^2$, wall-clock time to reach fit threshold, and expression complexity on the Nguyen benchmark --- note: we round to 1.0 for $R^2>0.99999$. The five leftmost methods were run by us under identical settings, seeds, and hardware. CaMo$^\dagger$ values are reported by \citet{liu2025camo} and obtained on different hardware and under CaMo's own data sampling. Its runtime is not reported and we draw no speed comparison against it.}
\label{tab:nguyen_runtime}
\footnotesize
\setlength{\tabcolsep}{5pt}
\begin{tabular}{@{}l rrrrr r@{}}
\toprule
\textbf{Problem} & \textbf{SymMatika} & \textbf{PySR} & \textbf{Operon} & \textbf{GP-GOMEA} & \textbf{SBP-GP} & \textbf{CaMo}$^\dagger$ \\
\midrule
\multicolumn{7}{@{}l}{\textbf{Test $R^2$}} \\
\cmidrule(r){1-1}\cmidrule(lr){2-6}\cmidrule(l){7-7}
Nguyen-1 & 1.0000 & 1.0000 & 1.0000 & 1.0000 & 1.0000 & 1.0000 \\
Nguyen-2 & 1.0000 & 1.0000 & 1.0000 & 1.0000 & 1.0000 & 1.0000 \\
Nguyen-3 & 1.0000 & 1.0000 & 1.0000 & 1.0000 & 1.0000 & 1.0000 \\
Nguyen-4 & 1.0000 & 1.0000 & 1.0000 & 1.0000 & 1.0000 & 0.9997 \\
Nguyen-5 & 1.0000 & 1.0000 & 1.0000 & 1.0000 & 1.0000 & 1.0000 \\
Nguyen-6 & 1.0000 & 1.0000 & 1.0000 & 1.0000 & 1.0000 & 0.9999 \\
Nguyen-7 & 1.0000 & 1.0000 & 1.0000 & 1.0000 & 1.0000 & 0.9956 \\
Nguyen-8 & 1.0000 & 1.0000 & 1.0000 & 1.0000 & 1.0000 & 1.0000 \\
Nguyen-9 & 1.0000 & 1.0000 & 1.0000 & 1.0000 & 1.0000 & 0.9991 \\
Nguyen-10 & 1.0000 & 1.0000 & 1.0000 & 1.0000 & 1.0000 & 0.9989 \\
Nguyen-11 & 1.0000 & 1.0000 & 1.0000 & 1.0000 & 1.0000 & 1.0000 \\
Nguyen-12 & 1.0000 & 1.0000 & 1.0000 & 0.9970 & 1.0000 & 0.9580 \\
\midrule
\textbf{Average} & 1.0000 & 1.0000 & 1.0000 & 0.9997 & 1.0000 & 0.9965 \\
\addlinespace[6pt]
\multicolumn{7}{@{}l}{\textbf{Time to fit threshold (s)}} \\
\cmidrule(r){1-1}\cmidrule(lr){2-6}\cmidrule(l){7-7}
Nguyen-1 & 0.34 & 34.41 & 1.24 & 6.50 & 1244.90 & --- \\
Nguyen-2 & 0.51 & 38.33 & 1.28 & 7.50 & 1507.90 & --- \\
Nguyen-3 & 2.30 & 37.25 & 1.31 & 6.30 & 1869.30 & --- \\
Nguyen-4 & 4.50 & 35.77 & 1.30 & 6.80 & 1783.60 & --- \\
Nguyen-5 & 0.38 & 33.16 & 1.25 & 11.40 & 1328.80 & --- \\
Nguyen-6 & 0.42 & 34.39 & 1.27 & 8.50 & 1184.50 & --- \\
Nguyen-7 & 4.32 & 36.58 & 1.32 & 8.00 & 1439.60 & --- \\
Nguyen-8 & 0.31 & 33.78 & 1.23 & 12.40 & 1187.50 & --- \\
Nguyen-9 & 1.42 & 34.23 & 1.21 & 9.30 & 1056.60 & --- \\
Nguyen-10 & 6.24 & 34.69 & 1.22 & 12.40 & 602.80 & --- \\
Nguyen-11 & 1.30 & 72.70 & 1.24 & 6.70 & 962.20 & --- \\
Nguyen-12 & 15.60 & 35.88 & 1.27 & 3.50 & 876.40 & --- \\
\midrule
\textbf{Average} & 3.14 & 38.43 & 1.26 & 8.28 & 1253.68 & --- \\
\addlinespace[6pt]
\multicolumn{7}{@{}l}{\textbf{Expression complexity}} \\
\cmidrule(r){1-1}\cmidrule(lr){2-6}\cmidrule(l){7-7}
Nguyen-1 & 9 & 13 & 22 & 18 & 389 & 14 \\
Nguyen-2 & 13 & 15 & 24 & 24 & 531 & 16 \\
Nguyen-3 & 17 & 17 & 24 & 26 & 391 & 23 \\
Nguyen-4 & 21 & 18 & 24 & 19 & 472 & 18 \\
Nguyen-5 & 9 & 9 & 24 & 18 & 313 & 16 \\
Nguyen-6 & 9 & 15 & 24 & 18 & 435 & 19 \\
Nguyen-7 & 10 & 18 & 24 & 24 & 463 & 19 \\
Nguyen-8 & 2 & 2 & 24 & 20 & 465 & 4 \\
Nguyen-9 & 6 & 17 & 23 & 28 & 470 & 12 \\
Nguyen-10 & 7 & 15 & 22 & 18 & 389 & 17 \\
Nguyen-11 & 3 & 7 & 24 & 17 & 415 & 5 \\
Nguyen-12 & 15 & 18 & 24 & 24 & 524 & 23 \\
\midrule
\textbf{Average} & 10.1 & 13.7 & 23.6 & 21.2 & 483.1 & 15.5 \\
\bottomrule
\end{tabular}
\end{table}

\begin{table}[H]
\centering
\caption{Time to fit $R^2>0.999$ threshold versus time to exact recovery on the Nguyen benchmarks (full model, 20 runs per problem)}
\label{tab:nguyen-runtime}
\begin{tabular}{l r r r}
\toprule
Problem & $t_{\mathrm{fit}}$ med.\ (s) & $t_{\mathrm{exact}}$ med.\ (s), recovered & Median $t$ (s), all runs \\
\midrule
Nguyen-1  & 0.17  & 0.17  & 0.17 \\
Nguyen-2  & 0.27  & 0.28  & 0.28 \\
Nguyen-3  & 0.32  & 6.10  & 7.19 \\
Nguyen-4  & 0.55  & 10.02 & 10.02 \\
Nguyen-5  & 0.56  & 0.65  & 1.09 \\
Nguyen-6  & 0.25  & 0.26  & 0.26 \\
Nguyen-7  & 0.71  & 1.37  & 2.01 \\
Nguyen-8  & 0.16  & 0.16  & 0.16 \\
Nguyen-9  & 0.69  & 0.69  & 0.69 \\
Nguyen-10 & 1.05  & 1.14  & 1.14 \\
Nguyen-11 & 0.65  & 0.64  & 0.65 \\
Nguyen-12 & 11.55 & 13.04 & 15.50 \\
\bottomrule
\end{tabular}
\end{table}

\begin{table}[H]
\centering
\caption{Per-run expressions returned by \textsc{SymMatika} on the two hardest Nguyen problems.
\cmark{} denotes symbolic equivalence to the target after simplification.
$\sin^{\circ 6}(y)$ denotes six nested applications of $\sin$.}
\label{tab:nguyen-per-run}
\begin{tabular}{lllc}
\toprule
Problem & Target & Returned expression & Exact \\
\midrule
\multirow{3}{*}{Nguyen-7}  & \multirow{3}{*}{$\log(x+1)+\log(x^2+1)$}
  & $\log(1+x+x^2+x^3)$ & \cmark \\
& & $\log\!\big(1+x^2+x^3+\log(e^{x})\big)$ & \cmark \\
& & $\log(1+x+x^2+x^3)$ & \cmark \\
\midrule
\multirow{3}{*}{Nguyen-12} & \multirow{3}{*}{$x^4-x^3+\tfrac{1}{2}y^2-y$}
  & $-x^3+x^4+0.5\,y^2-y$ & \cmark \\
& & $-x^3+x^4+0.5\,y^2-y$ & \cmark \\
& & $0.01-\sin\!\big(x^3-x^4+\sin^{\circ 6}(y)\big)$ & \xmark \\
\bottomrule
\end{tabular}
\end{table}

\begin{table}[H]
\centering
\caption{Cost of motif selection under each rule (default primitive set, 20 runs per cell).}
\label{tab:impact-cost}
\begin{tabular}{lcccccc}
\toprule
 & \multicolumn{3}{c}{Selection time (ms/gen)} & \multicolumn{3}{c}{Share of wall time (\%)} \\
\cmidrule(lr){2-4}\cmidrule(lr){5-7}
Problem & Random & Freq. & \textbf{Impact} & Random & Freq. & \textbf{Impact} \\
\midrule
Nguyen-4  & 0.54 & 0.48 & \textbf{0.26} & 3.4 & 3.2 & \textbf{1.6} \\
Nguyen-7  & 0.84 & 0.60 & \textbf{0.27} & 4.7 & 3.8 & \textbf{1.7} \\
Nguyen-12 & 1.16 & 1.61 & \textbf{0.61} & 3.0 & 3.8 & \textbf{1.6} \\
\bottomrule
\end{tabular}
\end{table}

\begin{table}[H]
    \centering
    \caption{Symbolic targets and example recovered expressions for the Eureqa implicit regression problems.}
    \begin{tabular}{l l l}
    \toprule
    \textbf{Problem} & \textbf{Target Equation} & \textbf{Recovered Expressions} \\
    \midrule
    \texttt{circle\_1} & $f = x^2 + y^2$ & 1. $f = x^2 + y^2$ \\
     & & 2. $f = x^2 + y^2$ \\
    \midrule
    \texttt{pendulum\_h\_1} & $f = \omega^2 - 19.6\cos(\theta)$ & 1. $f = \omega^2 - 19.6\cos(\theta)$ \\
     & & 2. $f = \frac{\theta^2}{\cos{\left(\omega\right)}}$ \\
    \midrule
    \texttt{sphere\_1} & $f = x^2 + y^2 + z^2$ & 1. $f = x^2 + y^2 + z^2$ \\
     & & 2. $f = x^2 + y^2 + z^2$ \\
    \midrule
    \texttt{dbl\_pendulum\_h\_1} & $f = \omega_1^2 + 0.5\omega_2^2 + \omega_1\omega_2\cos(\theta_1 - \theta_2)$ & 1. $f = \tan(4.8\cdot\tan(-6.1\cdot\omega_1 + \pi\cdot\omega_2))\cdot\theta_2$ \\
     & $- 19.6\cos(\theta_1) - 9.8\cos(\theta_2)$ & $\qquad -\theta_1$ \\
     & & 2. $f = \cos(-14.7\cdot\omega_1 + 9.4\cdot\omega_2 + \theta_1\cdot\sqrt\omega_2$ \\
     & & $\qquad + \sqrt{\theta_2\cdot\theta_1})$ \\
    \bottomrule
    \end{tabular}
    \label{tab:eureqa_targets}
\end{table}

To showcase some sample output, we run on a sample equation from the Feynman Lectures on Physics \citep{feynman2015feynman}, the relativistic Doppler shift formula $\omega = \frac{1+\frac{v}{c}}{\sqrt{1-\frac{v^2}{c^2}}}\omega_0$. After running, we return the expression $\omega=\sqrt{\frac{c+v}{\left(c - v\right)}}\omega_0$ with a MLE loss of $-3.10862\times 10^{-16}$, and the discovered expression is mathematically equivalent to the ground truth expression, albeit syntactically slightly different. We provide more outputs on a subset of sampled Feynman Equations in Appendix Sec.~\ref{sample-output}.

\subsection{Limitations}\label{lim}

\paragraph{Long expressions with few reusable substructures.} Demonstrated with I.12.11 and I.13.4 in Sec.~\ref{sec:motif_analysis}, the motif library can extract subtrees from partial solutions. Other Feynman equations do not decompose so neatly: II.35.18, $n = n_0 / \left(\exp(\mu_m B / k_b T) + \exp(-\mu_m B / k_b T)\right)$ has a single deeply nested term, and I.29.16, $x = \sqrt{x_1^2 + x_2^2 - 2x_1x_2\cos(\theta_1 - \theta_2)}$, mostly inside one unary argument. Such targets contain few easily separable subtrees, so capabilities on exact symbolic recovery on more complex problems is uncertain. Our Feynman results (Sec.~\ref{sec:feynman}) measure predictive accuracy and complexity, not exact recovery; measuring recovery across all 119 equations under the multi-seed protocol used for Nguyen is future work, and would directly test the library on long, composite expressions.

\paragraph{Rigorous testing of complex floating-point constants.} We performed limited evaluation on symbolic regression problems containing complex floating-point constants. We performed preliminary testing on the popular Nguyen-1$c$ problem $f=3.39x^3+2.12x^2+1.78x$, and report mean MLE of $-3.36 \times10^{-4}$ over 10 independent runs, with the common symbolic expression found being $f=3.4x^3+2.1x^2+1.8x$; this shows early evidence \textsc{SymMatika}'s capability to handle complex floating-point constants. However, these experiments were not rigorous and without adequate statistical testing. We intend to perform additional studies on similar problems to effectively evaluate \textsc{SymMatika} on this family of symbolic regression problems. Additionally, we will investigate alternative methods of constant optimization, as our current approach of scaling a constant $c$ by $s \sim \text{U}(0.5,1.5)$ has no explicit optimization signal and rounding to one-decimal precision constrains the system to solve symbolic regression problems with higher-precision constants, e.g. $3.39$ in Nguyen-1$c$.

\paragraph{Difficulty of handling complex variable interdependencies.} In our implicit regression experiments (Tab.~\ref{eureqa}), we observed consistent recovery on the first three problems, but difficulty with \texttt{dbl\_pendulum\_h\_1} (i.e. $f = \omega_1^2 + 0.5 \omega_2^2 + \omega_1 \omega_2 \cos(\theta_1 - \theta_2) - 19.6 \cos(\theta_1) - 9.8 \cos(\theta_2)$). We explain how \textsc{SymMatika} handles interdependent variables in Sec.~\ref{sec:problem_setup} and showcase results on \texttt{sphere\_1}, however \texttt{dbl\_pendulum\_h\_1} is a much more challenging problem. We aim to further investigate methods of handling more complex variable interdependencies in future work.

\begin{figure}[t]
\centering
\includegraphics[width=0.95\linewidth,height=0.32\textheight,keepaspectratio]{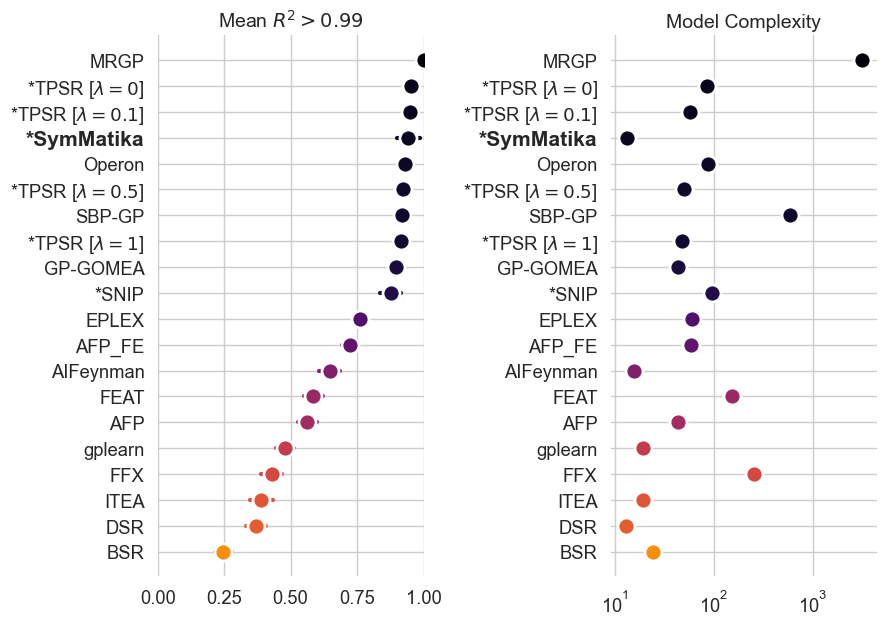}

\vspace{1em}

\includegraphics[width=0.95\linewidth,height=0.32\textheight,keepaspectratio]{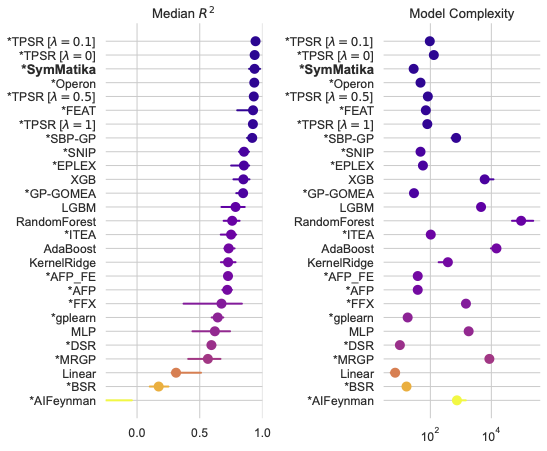}

\caption{95\% confidence intervals for (top) Feynman equations and (bottom) Black-box benchmarks. ``*'' refers to SR methods for Black-box datasets.}
\label{fig:feynman_bb}
\end{figure}

\begin{table*}[t]
\centering
\caption{Black-box vs. Feynman benchmark predictive accuracy and model complexity performances.}
\label{tab:sidebyside}

\begin{minipage}{0.48\textwidth}
\centering
\textbf{Black-box}\\[2pt]
\begin{tabular}{lcc}
\toprule
Algorithm & Median $R^2$ & Complexity \\
\midrule
TPSR ($\lambda{=}0.1$) & 0.945 & 95.710 \\
TPSR ($\lambda{=}0$)   & 0.938 & 129.850 \\
Operon                 & 0.938 & 61.742 \\
\textbf{SymMatika}     & 0.931 & 28.059 \\
TPSR ($\lambda{=}0.5$) & 0.931 & 82.580 \\
TPSR ($\lambda{=}1$)   & 0.924 & 79.430 \\
SBP-GP                 & 0.916 & 616.735 \\
FEAT                   & 0.906 & 72.664 \\
SNIP                   & 0.853 & 47.527 \\
GP-GOMEA               & 0.836 & 32.095 \\
EPLEX                  & 0.836 & 53.147 \\
XGB                    & 0.817 & 19133.300 \\
LGBM                   & 0.794 & 4822.224 \\
RandomForest           & 0.753 & 1155301.000 \\
AdaBoost               & 0.747 & 9892.822 \\
ITEA                   & 0.736 & 103.935 \\
KernelRidge            & 0.729 & 1170.091 \\
AFP                    & 0.710 & 36.262 \\
AFP\_FE                & 0.702 & 36.864 \\
FFX                    & 0.665 & 1429.411 \\
MLP                    & 0.638 & 1794.727 \\
gplearn                & 0.638 & 22.744 \\
DSR                    & 0.600 & 10.556 \\
MRGP                   & 0.502 & 9878.172 \\
Linear                 & 0.342 & 6.964 \\
BSR                    & 0.184 & 23.356 \\
AIFeynman              & -0.608 & 1989.177 \\
\bottomrule
\end{tabular}
\end{minipage}
\hfill
\begin{minipage}{0.48\textwidth}
\label{fig:feynman}
\centering
\textbf{Feynman}\\[2pt]
\begin{tabular}{lcc}
\toprule
Algorithm & Mean $R^2 > 0.99$ & Complexity \\
\midrule
MRGP                 & 0.977 & 3669.357 \\
TPSR ($\lambda{=}0$) & 0.952 & 84.420 \\
TPSR ($\lambda{=}0.1$) & 0.949 & 57.220 \\
\textbf{SymMatika}   & 0.942 & 13.294 \\
SBP-GP               & 0.941 & 489.416 \\
Operon               & 0.940 & 69.875 \\
SNIP                 & 0.882 & 31.630 \\
GP-GOMEA             & 0.880 & 34.571 \\
EPLEX                & 0.780 & 52.945 \\
AFP\_FE              & 0.733 & 39.966 \\
AIFeynman            & 0.685 & 124.478 \\
FEAT                 & 0.616 & 205.305 \\
AFP                  & 0.568 & 36.869 \\
gplearn              & 0.504 & 72.427 \\
FFX                  & 0.425 & 271.693 \\
ITEA                 & 0.405 & 21.098 \\
DSR                  & 0.358 & 14.857 \\
BSR                  & 0.222 & 25.500 \\
\bottomrule
\end{tabular}
\end{minipage}

\end{table*}

\end{document}